\documentclass{article}

\usepackage{amsmath, amssymb, amsfonts, mathtools}
\usepackage{physics}
\usepackage{cancel}
\usepackage{accents}
\usepackage{bm}

\makeatletter
\newcommand{\ostar}{\mathbin{\mathpalette\make@circled *}}

\newcommand{\make@circled}[2]{%
  \ooalign{$\m@th#1\smallbigcirc{#1}$\cr\hidewidth$\m@th#1#2$\hidewidth\cr}%
}
\newcommand{\smallbigcirc}[1]{%
  \vcenter{\hbox{\scalebox{0.77778}{$\m@th#1\bigcirc$}}}%
}
\makeatother

\usepackage{pict2e, picture}
\makeatletter
\DeclareRobustCommand{\Arrow}[1][]{%
\check@mathfonts
\if\relax\detokenize{#1}\relax
\settowidth{\dimen@}{$\m@th\rightarrow$}%
\else
\setlength{\dimen@}{#1}%
\fi
\sbox\z@{\usefont{U}{lasy}{m}{n}\symbol{41}}%
\begin{picture}(\dimen@,\ht\z@)
\roundcap
\put(\dimexpr\dimen@-.7\wd\z@,0){\usebox\z@}
\put(0,\fontdimen22\textfont2){\line(1,0){\dimen@}}
\end{picture}%
}
\makeatother

\usepackage{scalerel}
\DeclareMathAlphabet{\nummathbb}{U}{BOONDOX-ds}{m}{n}

\usepackage{algorithm}
\usepackage{algpseudocode}
\algtext*{EndWhile}
\algtext*{EndIf}
\algtext*{EndFor}
\algtext*{EndElse}

\makeatletter
\DeclareRobustCommand\widecheck[1]{{\mathpalette\@widecheck{#1}}}
\def\@widecheck#1#2{%
    \setbox\z@\hbox{\m@th$#1#2$}%
    \setbox\tw@\hbox{\m@th$#1%
       \widehat{%
          \vrule\@width\z@\@height\ht\z@
          \vrule\@height\z@\@width\wd\z@}$}%
    \dp\tw@-\ht\z@
    \@tempdima\ht\z@ \advance\@tempdima2\ht\tw@ \divide\@tempdima\thr@@
    \setbox\tw@\hbox{%
       \raise\@tempdima\hbox{\scalebox{1}[-1]{\lower\@tempdima\box
\tw@}}}%
    {\ooalign{\box\tw@ \cr \box\z@}}}
\makeatother

\usepackage{microtype}
\usepackage{graphicx}
\usepackage{booktabs} 

\usepackage{hyperref}

\usepackage[accepted]{icml2025}

\icmltitlerunning{Accepted at ICML 2025 Workshop on Efficient Systems for Foundational Models}

\usepackage{fancyhdr}

\usepackage{amsmath}
\usepackage{amssymb}
\usepackage{mathtools}
\usepackage{amsthm}

\usepackage[capitalize,noabbrev,nameinlink]{cleveref}

\theoremstyle{plain}

\theoremstyle{definition}

\theoremstyle{remark}

\usepackage[textsize=tiny]{todonotes}

\newcommand{\ourmethod}{\textsc{VocabTrim}}

\usepackage{multirow}

\usepackage{caption}
\usepackage{subcaption}

\usepackage[toc,page,header]{appendix}
\usepackage{minitoc}

\usepackage{enumitem} 

\usepackage{adjustbox}

\usepackage{xcolor}

\usepackage{optidef}

\usepackage{tikz}
\usetikzlibrary{matrix}

\usepackage{makecell}
\usepackage{color}

\usepackage{xargs}

\usepackage{pifont}

\usepackage[normalem]{ulem}

\begin{document}

\twocolumn[
\icmltitle{\ourmethod{}: Vocabulary Pruning for Efficient Speculative Decoding in LLMs}



\icmlsetsymbol{equal}{*}

\begin{icmlauthorlist}
\icmlauthor{Raghavv Goel}{comp}
\icmlauthor{Sudhanshu Agrawal}{comp}
\icmlauthor{Mukul Gagrani}{comp}
\icmlauthor{Junyoung Park}{comp}
\icmlauthor{Yifan Zao}{comp}
\icmlauthor{He Zhang}{comp}
\icmlauthor{Tian Liu}{comp}
\icmlauthor{Yiping Yang}{comp}
\icmlauthor{Xin Yuan}{comp}
\icmlauthor{Jiuyan Lu}{comp}
\icmlauthor{Chris Lott}{comp}
\icmlauthor{Mingu Lee}{comp}
\end{icmlauthorlist}

\icmlaffiliation{comp}{Qualcomm AI Research. Qualcomm AI Research is an initiative of Qualcomm Technologies, Inc.}

\icmlcorrespondingauthor{Raghavv Goel, Mukul Gagrani, Mingu Lee}{raghgoel/mgagrani/mingul@qti.qualcomm.com}

\icmlkeywords{Machine Learning, ICML}

\vskip 0.3in
]
\printAffiliationsAndNotice{}
\begin{abstract}

In this paper, we introduce a simple training-free technique to improve the performance of drafter-based speculative decoding (SpD) methods that incorporates language modeling head (LM head) during drafting process. A drafter-based speculative decoding leverages one or more smaller language models, a.k.a. {\it drafters} or {\it draft models}, to sample a draft sequence or tree consisting of multiple tokens, followed by verification by a base LLM, a {\it target model}, accepting a subset as its valid generation. As it is usually considered that the speculative decoding requires one-to-one mapping between vocabularies of the target model and the draft model, it has been natural to share the vocabulary between them, or even share the LM head as in EAGLE or Medusa. We first identify that this draft token sampling scheme inherently contains an unnecessary inference overhead in drafting, especially for some target LLMs with very large vocabularies.  Then, we propose a simple technique, \ourmethod{}, to mitigate the drafting overhead to improve the generation speed in memory-bound environment. \ourmethod{} reconstructs the drafter LM head to contain only a limited set of tokens, selected by the most frequently sampled from the vocabulary of the target model. While limiting the vocabulary in drafting slightly degrades the acceptance rate, it significantly reduces the drafting latency in memory-bound process which is often the case on edge devices, resulting in higher memory-bound speed up (MBSU). We show that our method can boost the memory-bound speed-up for Llama-3 models on Spec-Bench, specifically by $16\%$ for Llama-3.2-3B-Instruct.

\end{abstract}

\section{Introduction}
\label{sec:introduction}
\begin{figure}[]
    \centering
    \includegraphics[width=\linewidth]{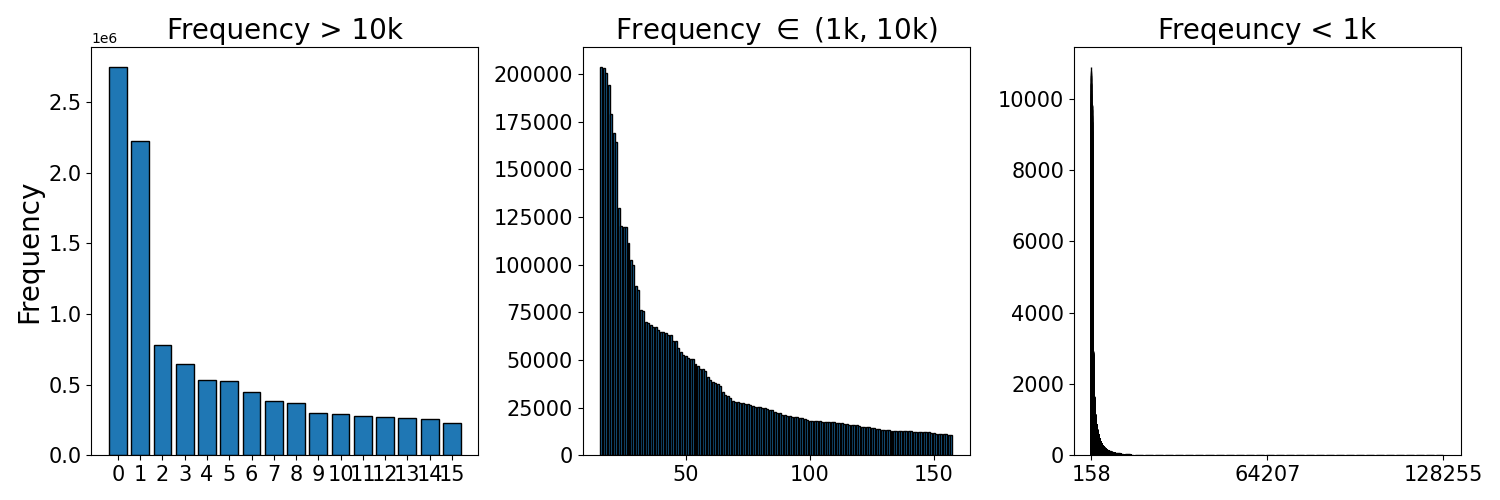}
    \caption{xLAM (function-calling) dataset token freqeuncy based on target model generation. Plots divided into three parts for ease of readability.}
    \label{fig:xlam-bar-plot}
\vspace{-0.5cm}
\end{figure}

Speculative Decoding (SpD) \cite{leviathan2023fast} is a widely adopted inference optimization technique for large language models (LLMs). In SpD, a lightweight drafter speculates the tokens that a larger target model would generate. The target model then selectively accepts or rejects these speculations based on a policy that often ensures that the output follows the same distribution as the target model. Prior work on SpD has focused primarily on balancing the expressiveness and efficiency of the drafter, either by designing novel model architectures \cite{li2024eagle, cai2024medusa, zimmer2024mixture} or by developing draft-time search algorithms \cite{miao2023specinfer, jeon2024recursive}. However, these approaches typically require training a separate drafter, with the shared tokenizer, when a well-aligned pretrained model is not available \cite{goel2024direct}.

While recent advances have proposed increasingly effective drafters, we observe that their predictions tend to focus on “easy-to-predict” tokens such as articles, prepositions, or completions of partially generated words. For example, a drafter may suggest \texttt{table} immediately after the target model generates \texttt{vege}, thereby completing the word \texttt{vegetable} \cite{gagrani2024speculative}. This behavior suggests that the role of the drafter could be shifted from a general-purpose generator to a more specialized token suggester, particularly by restricting its vocabulary.

We further observe that in many downstream tasks, the target model generation is limited on a small portion of its full vocabulary. As shown in \cref{fig:xlam-bar-plot}, we analyze the output token distribution of Llama-3.2-3B-Instruct on the training split of a function calling dataset \cite{liu2024apigen}. Only a small set of tokens occur at very high frequencies. For instance, 15 tokens are sampled more than 10K times and the next 140 tokens appear between 1K and 10K times. In contrast, more than 120,000 tokens are rarely or not sampled at all.

This observation suggests an opportunity to simplify the output token space of the drafter. If the drafter only needs to predict a limited set of frequently sampled tokens, computing logits over the full vocabulary may be unnecessary, i.e., saving memory and computation for drafting. The saving becomes more significant when the target model is with a large vocabulary size, which is common in modern LLMs for improved support for multiple language and token efficiency, i.e., compression rate. \cite{dubey2024llama}
The language modeling (LM) head, which maps hidden representations to vocabulary logits through a linear projection, is often a major contributor to both model size and inference latency. This issue is particularly pronounced in small drafters with large vocabularies. For example, in a 314M-parameter drafter using the Llama 3 vocabulary (128K tokens), the LM head alone accounts for over 30\% of the total parameters. This significantly limits the speedup potential of SpD, given that generation is typically a memory-bound process.

Motivated by these observations, we propose \ourmethod{}, a training-free method for improving the efficiency of speculative decoding by reducing the size of the LM head of the drafter. 
By restricting the output dimension of the LM head, \ourmethod{} provides significant memory and latency savings without requiring any training or architectural changes. For Llama 3 models, we show that \ourmethod{} can reduce the output dimension of the LM heads by up to 75\% with negligible impact to the acceptance rates. When applied to the state-of-the-art SpD method EAGLE-2, our method achieves on average 16\% latency improvement on Spec-Bench tasks \cite{xia-etal-2024-unlocking}. 
We hope that this work encourages further research into optimizing previously underexplored components of the SpD pipeline.

\section{Related Work}
\label{sec:related_work}

Speculative decoding was introduced in \cite{leviathan2023fast}, which showed using two language models: a smaller drafter model, and the larger target model, in a memory bound setting can help accelerate LLM token generation. A key contribution of this work was the generations are lossless, i.e., they follow the target model distribution. Several works build on top of this, extending SpD to recursive speculative decoding \cite{jeon2024recursive, miao2023specinfer}, instead of training independent small drafters, extra LM-head were augmented \cite{cai2024medusa}, draft model was attached to last layer of LLM \cite{li2024eagle, zimmer2024mixture}, etc. Additionally, some work propose training separate drafters for different downstream tasks to achieve further boost in speed \cite{kimunified}, training drafters is cumbersome and resource intense as many times small models are simply unavailable \cite{goel2024direct}. We propose a plug-and-play, training-free method, based on the limited representation capacity of drafter model, and exploit for reducing computation while maintaining block efficiency with boost in memory-bound speed-up. 

Earlier work in \cite{gagrani2024speculative} has shown in multi-modal setting, that even using only a text-based drafter can achieve good speed-ups and their qualitative analysis shows that drafter primarily predicts easy words such as articles, prepositions, or completing half-generated words. 

Recently, AdaptiVocab \cite{nakash2025adaptivocab} explored using custom vocabularies for LLM generation, to reduce computation and memory cost for target (or base) model. A concept similar to our approach was independently proposed in \cite{zhao2025fr}, although their method is limited to pruning draft LM-head using only raw-text data. Detailed similarities and differences with \cite{zhao2025fr} are discussed in \cref{subsec:fr_spec}.
It is important to note that the LM head linear layer consumes a significant portion of memory and computation during inference. In contrast to AdaptiVocab, our work focuses  specifically on optimizing the LM head of drafter. This design choice ensures that the final performance of the target model remains unaffected, while still leveraging the lossless nature of speculative decoding methods.

\section{Method}
\label{sec:method}
\begin{figure}
    \centering
    \includegraphics[width=0.9\linewidth]{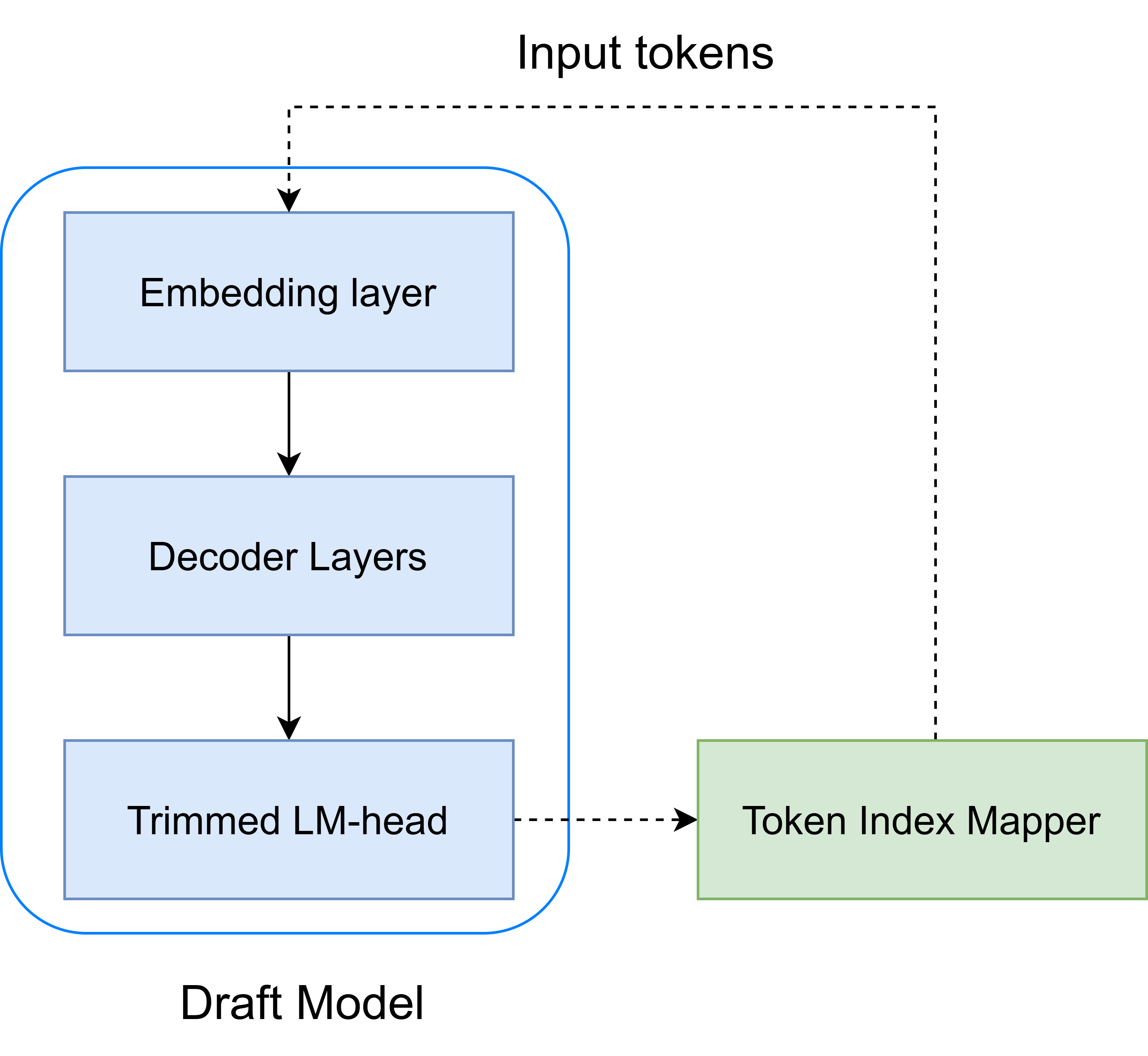}
    \caption{SpD inference with trimmed vocabulary of drafter}
    \label{fig:main-diagram}
\vspace{-0.5cm}
\end{figure}
Based on our observation that only a small subset of tokens occur dominantly in many language modeling tasks (as shown in \cref{fig:xlam-bar-plot}), we propose \ourmethod{}, a training-free and efficient Speculative Decoding (SpD) method that removes infrequent tokens from the draft model’s vocabulary. \ourmethod{} introduces minimal changes to the SpD pipeline and imposes no architectural constraints, facilitating seamless integration into existing SpD techniques, general flow is shown in \cref{fig:main-diagram} with algorithm in \cref{subsec:appendix_algo}.

For a draft model with vocabulary \( \mathbb{V} \) and LM head parameter \( W \), \ourmethod{} constructs a trimmed vocabulary \( \mathbb{V}^{\text{Trim}} \) and a corresponding trimmed LM head \( W^{\text{Trim}} \) by running the target model on a calibration dataset \( \mathcal{D} \) and selecting the most frequently occurring tokens in \( \mathbb{V} \) along with their corresponding rows in \( W \). Formally, vocabulary trimming is defined as follows:
\begin{equation}
    \begin{aligned}
        \mathbb{V}^{\text{Trim}} &= \mathbb{V}[\textrm{Top-K}(c,k)] \\
        W^{\text{Trim}} &= W[\textrm{Top-K}(c,k), :] \\
    \end{aligned}
\end{equation}
where \( c \) is the token frequency counter and \( k \) is the desired size of the trimmed vocabulary. The frequency counter \( c \) is computed by counting how often each token in \( \mathbb{V} \) appears across the calibration dataset \( \mathcal{D} \), as shown in \cref{algorithm:token_counting}.

Two types of calibration datasets are considered: (a) raw text data, and (b) target model generation. Raw text data is available in ample amounts, even for many evaluation tasks which have train splits. Target model generation may also be readily available as it is used to finetune the drafter \cite{goel2024direct}. If not, it may be easily generated by simply letting the target model to generate completions following queries from a dataset. As an ablation, we additionally use drafter generated data (used for finetuning drafter in \cite{zhou2023distillspec}). We find that the target model generated calibration dataset performs the best (least drop in acceptance rate while maximum increase in memory-bound speed-up) as shown in \cref{tab:specbench_llama3b_table}.

Note that, in this paper, we restrict \ourmethod{} experiments with Top-$K$-based trimming of the drafter LM-head. However, our framework is general and also supports choosing based on Top-$P$, or based on minimum frequency of tokens in calibration datasets.
Moreover, other selection criteria for trimmed vocabulary  can be: compute resources available, hardware-memory constraints, or maximum allowable drop in accuracy.
We leave further exploration including these as a future work.


\begin{algorithm}[t]
\caption{Count Token Frequencies}
\begin{algorithmic}[1]
\Require Calibration dataset \( \mathcal{D} \)
\State Initialize counter vector \( c \in \mathbb{N}^{|\mathbb{V}|} \) with zeros
\For{each \( x \in \mathcal{D} \)}
    \State \( [t_0, t_1, \dots, t_n] \leftarrow \text{Tokenize}(x) \)
    \For{\( i = 0 \) to \( n \)}
        \State \( c[t_i] \leftarrow c[t_i] + 1 \)
    \EndFor
\EndFor
\State \textbf{Return} \( c \)
\end{algorithmic}
\label{algorithm:token_counting}
\end{algorithm}
\vspace{-0.5cm}

\section{Experiments}
\label{sec:experiments}
To show the efficacy of such trimmed-vocabulary drafter models, we perform experiments using the L{\small LAMA}3 models  \cite{grattafiori2024llama}, note that we perform greedy decoding for target model which implies a token is only accepted if exactly matched with the target model, enabling lossless generations. 
In the following experiments, we assume a fixed $K$ depending on the evaluation task. We additionally ablate over different sizes of draft LM head ($W^{\text{Trim}}$) to compare the effect on acceptance and speed-up increase. Our experiment setup and other details are as follows:

\subsection{Settings}
\textbf{Models}: We consider two Llama 3 models of different sizes: Llama-3.2-3B-Instruct and Llama-3.1-8B-Instruct. For each target model, we use two kinds of drafter architectures (a) EAGLE-based SpD \cite{li2024eagle2}, and (b) independent drafter-based SpD \cite{leviathan2023fast}. However, in both cases, we use the same draft tokens sampling, draft trees construction, and verification following EAGLE-2.

\textbf{Drafter Training}: For (a), the EAGLE drafter model is trained following EAGLE \cite{li2024eagle}, while, for (b), we follow \cite{goel2024direct} to train a $314$M standalone drafter (detailed configuration mentioned in \cref{sec:appendix_independent_dm}).

\textbf{Tasks}: We use tasks from Speculative-Decoding benchmark (Spec-Bench) \cite{xia-etal-2024-unlocking} which covers several tasks including summarization, coding, and math tasks (from GSM8K \cite{cobbe2021gsm8k}) as well as additional tasks such as function calling (xLAM) \cite{liu2024apigen}, and open-ended text generation (creative writing subset of Dolly-15k) \cite{DatabricksBlog2023DollyV2}.

\textbf{Performance Metrics}: (1) \textbf{block efficiency} ($\tau$): average number of tokens generated per block (or target model run), for a block of size $\gamma$ and input $x$, the maximum value of $\tau(x)$ can be $\gamma + 1$, (2) memory-bound speed-up (\textbf{MBSU}): theoretical speed-up achieved by SpD for a given block efficiency $\tau(x)$ and a relative latency $c$ defined as ratio between number of parameters of draft to target model, i.e., $\mathrm{MBSU}(x):=\frac{\tau(x)}{c\gamma + 1}$,

\textbf{Calibration Datasets}: 
For general text generation tasks such as open-ended text generation, summarization, etc., we use samples from Minipile \cite{kaddour2023minipile} as the raw dataset calibration. For target model-generated calibration, we use instruction-based dataset: OIG-small-chip2 \footnote{\href{https://huggingface.co/datasets/0-hero/OIG-small-chip2}{OIG-small-chip2 huggingface link}} and OpenAssistant \footnote{\href{https://huggingface.co/datasets/OpenAssistant/oasst1}{OpenAssistant huggingface link}} for Llama3.2-3B-Instruct and Llama3.1-8B-Instruct respectively. OIG is also used to generate drafter based  calibration dataset. Lastly, for function-calling evaluation task (xLAM), samples from train split are used for raw-dataset, target generations, and draft generation based calibration datasets.

\textbf{Draft Tree}: 
We use the draft tree with the depth of $3$ with top-K$=8$ and maximum tokens $=32$ for all generation of both EAGLE drafter and independent drafter. 

\begin{table*}[]
    \caption{Spec-Dec Benchmark with Eagle and Independent-drafter based SpD for Llama3.2-3B-Instruct}
    \label{tab:specbench_llama3b_table}
    \begin{center}
    \resizebox{\textwidth}{!}{%
    \begin{tabular}{lccccccccccccccccccc}
    \toprule
   Method  & LM$_{\text{Head},\text{Draft}}$ & \multicolumn{2}{c}{Writing} & \multicolumn{2}{c}{Roleplay} &  \multicolumn{2}{c}{Math/Reasoning} & \multicolumn{2}{c}{Coding} & \multicolumn{2}{c}{Extraction} & \multicolumn{2}{c}{Translation} & \multicolumn{2}{c}{Summarization} & \multicolumn{2}{c}{QA} & \multicolumn{2}{c}{RAG}
    \\
     \cmidrule[0.8pt](lr){3-20}
    & (M) & BE & MBSU & BE & MBSU & BE & MBSU & BE & MBSU & BE & MBSU & BE & MBSU & BE & MBSU & BE & MBSU & BE & MBSU 
    \\
   \cmidrule[0.8pt](lr){1-20}
   Eagle  & 394.0 & 3.141 & 1.475 & 3.092 & 1.451 & 3.494 & 1.64 & 3.638 & 1.708 & 3.693 & 1.734 & 3.3 & 1.549 & 3.098 & 1.454 & 3.214 & 1.509 & 3.23 & 1.516 
   \\
   \cmidrule[0.8pt](lr){2-20}
   +Raw-dataset & 101.3 & 2.974 & 1.685 & 3.062 & \textbf{1.735} & 3.421 & 1.938 & 3.294 & 1.866 &  3.405 & 1.928 & 3.14 & 1.778 & 2.996 & \textbf{1.697} & 3.098 & 1.755 & 3.131 & 1.774
   \\
   +Target generated & 101.3 & 3.081 & \textbf{1.745} & 3.038 & 1.721 & 3.443 & \textbf{1.950} & 3.434 & \textbf{1.945} & 3.588 & \textbf{2.032} & 3.241 & \textbf{1.836} & 2.993 & 1.695 & 3.165 & \textbf{1.793} & 3.133 & \textbf{1.775} 
   \\
   +Draft generated & 101.3 & 3.057 & 1.732 & 3.016 & 1.708 & 3.43 & 1.943 & 3.379 & 1.914 & 3.518 & 1.993 & 3.229 & 1.829 & 2.98 & 1.688 & 3.148 & 1.783 & 3.073 & 1.741
   \\
   \bottomrule
   \bottomrule
   Independent  & 131.3 & 3.846 & 2.765 & 3.864 & 2.778 & 4.434 & 3.187 & 3.999 & 2.875 & 4.780 & 3.436 & 4.221 & 3.035 & 3.239 & 2.328 & 4.160 & 2.991 & 3.943 & 2.835 
   \\
   \cmidrule[0.8pt](lr){2-20}
   +Raw-dataset & 33.8 & 3.682 & 2.900 & 3.551 & 2.797 & 4.224 & 3.327 & 3.592 & 2.829 & 4.332 & 3.412 & 3.975 & 3.131 & 3.097 & 2.439 & 3.883 & 3.059 & 3.737 & 2.944
   \\
   +Target generated  & 33.8 & 3.941 & \textbf{3.104} & 3.781 & \textbf{2.978} & 4.379 & \textbf{3.449} & 3.707 & \textbf{2.920} & 4.704 & \textbf{3.705} & 4.147 & \textbf{3.267} & 3.086 & \textbf{2.431} & 4.078 & \textbf{3.212} & 3.769 & \textbf{2.969}
   \\ 
   +Draft generated  & 33.8 & 3.905 & 3.076 & 3.738 & 2.944 & 3.948 & 2.838 & 3.534 & 2.783 & 4.636 & 3.652 & 4.117 & 3.243 & 3.037 & 2.392 & 4.068 & 3.204 & 3.685 & 2.902
   \\ 
   \bottomrule
    \end{tabular}%
}
\end{center}    
\end{table*}

\begin{table}[]
    \caption{Eagle based SpD for Llama3.2-3B-Instruct model on open-ended text generation (DOLLY) and function calling (xLAM) task}
    \label{tab:eagle_llama3.2-3b-dolly-xlam}
    \begin{center}
    \resizebox{0.5\textwidth}{!}{%
    \begin{tabular}{lcccccc}
    \toprule
   Method  & \multicolumn{3}{c}{DOLLY} & \multicolumn{3}{c}{xLAM} 
    \\
     \cmidrule[0.8pt](lr){2-4} \cmidrule[0.8pt](lr){5-7} 
    & LM$_{\text{Head},\text{Draft}}$ (M) & BE & MBSU & LM$_{\text{Head},\text{Draft}}$ (M) & BE & MBSU
    \\
    \cmidrule[0.8pt](lr){1-7}
   Eagle  & 394.0 & 3.237 & 1.52 & 394.0 & 2.937 & 1.379 
   \\
   \cmidrule[0.8pt](lr){2-7}
   +Raw-dataset & 101.3 & 3.091 & 1.751 & 15.4 & 2.397 & 1.445 
   \\
   +Target generated & 101.3 & 3.193 & \textbf{1.809} & 15.4 & 2.860 & \textbf{1.725}
   \\
   +Draft generated & 101.3 & 3.173 & 1.797 & 15.4 & 2.777 & 1.674
   \\
   \bottomrule
   \bottomrule
   Independent  & 131.3 &  3.675 & 2.642 & 131.3 & 2.364 & 1.700
   \\
   \cmidrule[0.8pt](lr){2-7}
   +Raw-dataset & 33.8 &  3.483 & 2.744 & 5.1 & 2.075 & 1.681
   \\
   +Target generated & 33.8  & 3.629 & \textbf{2.858} & 5.1 & 2.344 & \textbf{1.900}
   \\
   +Draft generated & 33.8 & 3.589 & 2.827 & 5.1 & 2.314 & 1.875
   \\
   \bottomrule
    \end{tabular}
    }
\end{center}
\end{table}

\subsection{Results}
Llama-3.2-3B-Instruct results with \ourmethod{} on Spec-Bench are shown in \cref{tab:specbench_llama3b_table} for both the EAGLE drafter and the independent drafter. We observe that target-generated calibration dataset-based vocabulary trimming leads to the smallest drop in block efficiency ($2$-$5\%$) and the largest gains in MBSU ($14$-$18\%$), followed by the cases of using draft model generation and the raw text. 
With the independent drafter, in the similar vein, \ourmethod{} with target-generated calibration dataset outperforms the raw text and draft-generated dataset. The block-efficiency drop for target model generated calibration dataset is $1$-$7\%$ with the MBSU gain of $2$-$12.3\%$.

To study the performance of \ourmethod{} on other domains, we additionally conduct experiments on open-ended text generation with relatively long generation, and on a function calling task, for both the EAGLE and independent drafter \cref{tab:eagle_llama3.2-3b-dolly-xlam}. 
Note that, while the same trimmed vocabulary set can be shared across multiple tasks with similar nature of text (e.g. chat, QA, writing), it may degrade performance on some other tasks with little overlaps in expected target model generation (e.g., coding task \cref{subsec:appendix_limitations}). As such, on xLAM function call task, we use a different trimmed vocabulary with the size of $5K$ based on training-split of xLAM dataset, unlike the trimmed vocabulary used for Dolly with the size of $\sim 32K$ (same as the one used for Spec-Bench).
Similar to the previous result, using target-generated calibration dataset gives the highest boost in MBSU with the lowest drop in BE. For the EAGLE drafter, BE drops only $1.4\%$ with the MBSU gain of $19\%$ on Dolly, and $2.6\%$ BE drop on xLAM with MBSU gains of $25\%$. For the independent drafter, using target-generated calibration dataset drops BE by $1.3\%$ and $0.8\%$ with the MBSU gains of $\sim8\%$ and $11.8\%$ on Dolly and xLAM, respectively. 

\begin{table*}[]
    \caption{Spec-Dec Benchmark with Eagle based SpD for Llama3.1-8B-Instruct}
    \label{tab:eagle_specbench_llama8b}
    \begin{center}
    \resizebox{\textwidth}{!}{%
    \begin{tabular}{lccccccccccccccccccc}
    \toprule
   Method  & LM$_{\text{Head},\text{Draft}}$ & \multicolumn{2}{c}{Writing} & \multicolumn{2}{c}{Roleplay} &  \multicolumn{2}{c}{Math/Reasoning} & \multicolumn{2}{c}{Coding} & \multicolumn{2}{c}{Extraction} & \multicolumn{2}{c}{Translation} & \multicolumn{2}{c}{Summarization} & \multicolumn{2}{c}{QA} & \multicolumn{2}{c}{RAG}
    \\
     \cmidrule[0.8pt](lr){3-20}
    & (M) & BE & MBSU & BE & MBSU & BE & MBSU & BE & MBSU & BE & MBSU & BE & MBSU & BE & MBSU & BE & MBSU & BE & MBSU 
    \\
    \cmidrule[0.8pt](lr){1-20}
   Eagle  & 525.3 & 3.134 & 1.910 & 2.864 & 1.745 & 3.300 & 2.011 & 3.628 & 2.211 & 3.517 & 2.144 & 3.072 & 1.872 & 2.970 & 1.810 & 2.928 & 1.785 & 2.962 & 1.805 
   \\
   \cmidrule[0.8pt](lr){2-20}
   +Raw-dataset & 135.0 & 2.942 & 2.034 & 2.728 & 1.886 & 3.237 & 2.238 & 3.264 & 2.257 & 3.301 & 2.282 & 2.984 & 2.063 & 2.857 & 1.975 & 2.847 & 1.968 & 2.833 & 1.959
   \\
   +Target generated & 135.0 & 3.096 & \textbf{2.140} & 2.737 & \textbf{1.892} & 3.323 & \textbf{2.297} & 3.484 & \textbf{2.408} & 3.494 & \textbf{2.416} & 3.040 & \textbf{2.102} & 2.877 & \textbf{1.989} & 2.908 & \textbf{2.010} & 2.868 & \textbf{1.983}
   \\
   \bottomrule
    \end{tabular}%
}
\end{center}    
\end{table*}

For Llama-3.1-8B-Instruct, we perform experiments on Spec-Bench using the EAGLE drafter with the raw and the target-generated calibration dataset \cref{tab:eagle_specbench_llama8b}. Note that as the base model size is larger, the overall MBSU increases as the ratio of drafter size and target model size decreases. We observe that, even in this scenario, reduction in drafter size via smaller LM head improves MBSU ($8-12\%$ on Spec-Bench) with minimal BE ($1-4\%$ on Spec-Bench) degradation with target-generated calibration dataset outperforming the others.


\subsection{Ablation Study}
\subsubsection{Performance of EAGLE-2 with Various Drafter LM Head Sizes}
\begin{figure}
    \centering
    \includegraphics[width=1.0\linewidth]{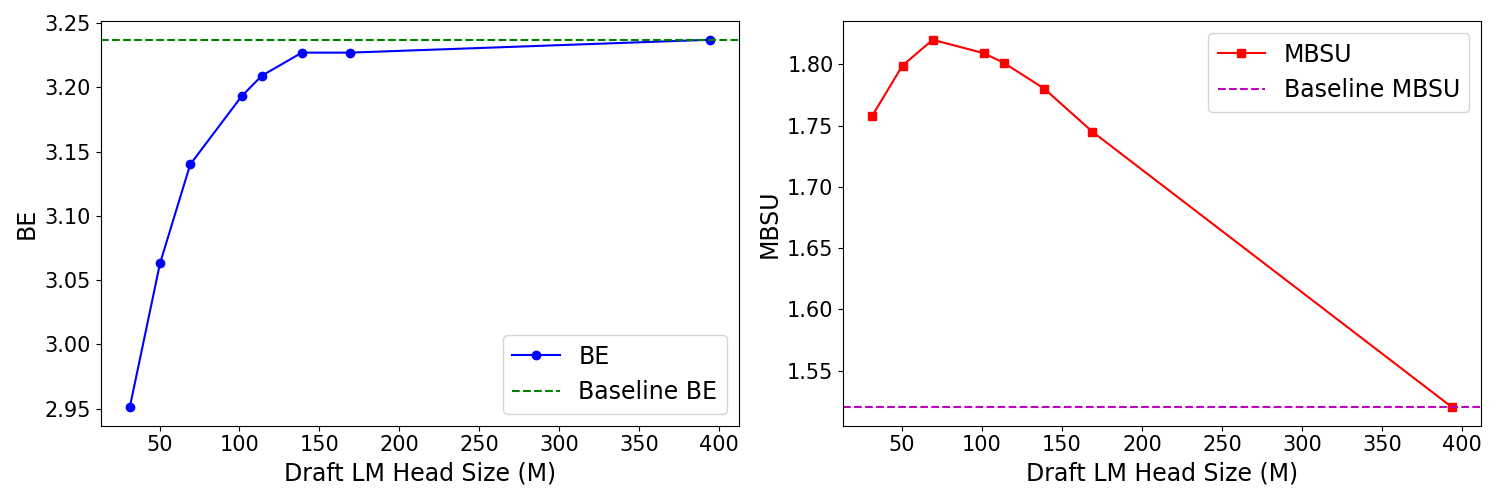}
    \caption{Llama3.2-3B-Instruct performance (BE, MBSU) with different draft LM head sizes.}
    \label{fig:dolly-ablation-llama3.2-3b-eagle}
\end{figure}

\begin{figure}
    \centering
    \includegraphics[width=1.0\linewidth]{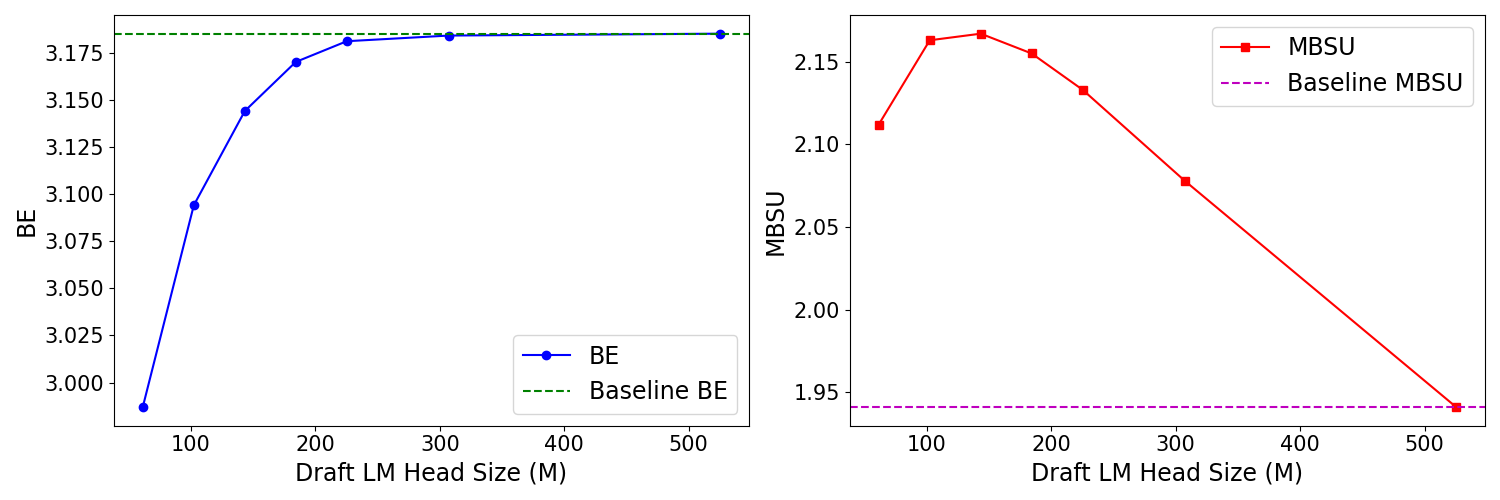}
    \caption{Llama3.1-8B-Instruct performance (BE, MBSU) with different draft LM head sizes.}
    \label{fig:dolly-ablation-llama3.1-8b-eagle}
\end{figure}

In \cref{fig:dolly-ablation-llama3.2-3b-eagle} and \ref{fig:dolly-ablation-llama3.1-8b-eagle}, we show the effects of amount of vocab trimming, i.e., draft LM head sizes, to BE and MBSU on Dolly dataset. As expected, BE increases accordingly as the draft LM head size increases. However, due to the trade-off between drafting cost and improved BE, the MBSU has a sweet spot at around $70$M sized $W^{\text{Trim}}$  corresponding to $k\sim 23K$ out of $128K$ tokens for Llama-3.2-3B-Instruct, boosting MBSU by $19.7\%$ with only a $\sim3\%$ drop in BE. Similarly, for Llama-3.1-8B-Instruct target, the highest MBSU is achieved by $143.4M$ sized draft LM-head consisting of $35K$ out of $128K$ tokens leading to only $1.2\%$ drop in block-efficiency with $11.6\%$ improvement in MBSU.

\subsubsection{Performance of EAGLE-2 With Various Draft Tree Hyperparameters}

\begin{table}[]
    \caption{Eagle based SpD for Llama3.2-3B-Instruct model on open-ended text generation (DOLLY) for different tree depth with and without draft model vocabulary trimming}
    \label{tab:eagle_decoding_tree_llama3.2-3b-dolly}
    \begin{center}
    \resizebox{0.45\textwidth}{!}{%
    \begin{tabular}{lcccc}
    \toprule
   Depth  & \multicolumn{2}{c}{Baseline} & \multicolumn{2}{c}{Baseline+\ourmethod{}} 
    \\
     \cmidrule[0.8pt](lr){2-3} \cmidrule[0.8pt](lr){4-5} 
     & BE & MBSU & BE & MBSU
    \\
    \cmidrule[0.8pt](lr){1-5}
   5  & 3.551 & 1.318 & 3.504 & 1.631
   \\
   6 & 3.578 & 1.202 & 3.510 & 1.500
   \\
   7 & 3.567 & 1.094 & 3.510 & 1.387
   \\
    \bottomrule
    \end{tabular}
    }
\end{center}
\end{table}

We ablate over different tree depth using EAGLE-2 decoder to gauge its impact on \ourmethod{}, with total draft tokens $=60$. We use top-$K$ $=32$k for trimmed drafter LM head as used in previous results. The original size of LM-head is $394$M while the size of trimmed LM-head is $101.3$M. The results are shown in \cref{tab:eagle_decoding_tree_llama3.2-3b-dolly}, where we observe that for different tree-depths, \ourmethod{} gives MBSU boost with minimal drop in performance, the absolute of MBSU increase remains consistent. 

\section{Conclusion}
We propose a new direction of improving performance of drafter-based speculative decoding methods by reducing the size of drafter LM head. Our method \ourmethod{} is a training-free off-the-shelf technique that can be easily integrated with various speculative decoding methods that incorporates LM head layers. Experiments with Llama 3 models on various downstream task shows that drafters can operate on smaller vocabulary space giving significant boost in performance in memory bound token generation process of LLMs. 

\section{Acknowledgment}
We thank Weilin Zhao and Xu Han who kindly pointed out the resemblance of the idea in our paper to their prior work \cite{zhao2025fr}. We discuss their similarities and differences in \cref{sec:related_work} and \cref{subsec:fr_spec}.
\bibliography{reference}
\bibliographystyle{icml2025}

\appendix
\section{Discussion}

\subsection{Future Work}
\label{subsec:appendix_future}
In current work, we rely on limited number of datasets for different tasks, specifically, Minipile \cite{kaddour2023minipile} for computing occurrence of tokens in raw-dataset, and OIG-small-chip2, OpenAssistant datasets for generating target (and draft) model dataset, and xLAM \cite{liu2024apigen} train-split for function-calling task. In following works, we will experiment specific downstream tasks using more task-specific domain dataset, for example, using train-split from GSM8k \cite{cobbe2021gsm8k} for math tasks, and using train split from HumanEval \cite{chen2021evaluating} for coding tasks, we believe this will further reduce the acceptance-rate gap while boosting MBSU. 
The idea of having separate draft model LM-head for separate downstream task can be considered analogous to low-rank adaptation (LORA) for LLMs \cite{hu2022lora}, where for each downstream task, a set of LORA parameters are trained. 


\subsection{Limitations}
\label{subsec:appendix_limitations}
We observe that the coding task in \cref{tab:specbench_llama3b_table} leads to slightly higher BE drop ($5.6\%$ for Eagle-drafter using target based calibration dataset) than other tasks with lower MBSU gains ($13.9\%$). We believe this is due to mismatch in the token distribution used in code-generation compared to the token distribution used in our calibration datasets (general english). Using a code-dataset for calibrating the VocabTrim would be beneficial in this scenario, to further boost the performance.

\subsection{Comparison of \ourmethod{} and FR-Spec: Similarities and Differences}
\label{subsec:fr_spec}
It has come to our attention that \ourmethod{} and FR-Spec \cite{zhao2025fr} share the idea of pruning LM head for speculative decoding. In this section, we highlight some key differences between them. 
First, they commonly identify the LM head as a bottleneck in draft model as motivation for the proposed works. While the induced algorithms resemble to each other, we emphasize that our focus is on the memory bandwidth of drafting process associated with the size of the drafter, regardless of its computational complexity. This is crucial in memory-bound environment, which is often the case in edge devices but not obvious in larger scale devices such as Nvidia GPUs in which cases the methods may lead to only marginal improvements. In this context, we report MBSU (memory-bound speed up) as the primary performance metric and make the performance measurement more consistent with on-edge-device performance.

Second, we propose the use of model-generated datasets, i.e., distillation datasets, to construct the token distribution, under the motivation of distribution matching and leading to the best performance on many cases, while comparing different types of calibration datasets: (a) target-generated, (b) draft-generated, and (c) raw dataset, similarly in \cite{zhao2025fr}.

Lastly, our approach is a generic method for speculative decoding method with drafter LM head in the loop, not limited to EAGLE. Thus, we compare both independent drafter model and EAGLE-style drafter model.

\section{Independent Draft Model}
\label{sec:appendix_independent_dm}
\subsection{Model configurations}
\label{subsec:model_config}
The following configurations are used for $314$M draft models following \cite{biderman2023pythia}:
\begin{table}[h]
    \centering
    \caption{Draft model configurations}
    \begin{tabular}{l||r}
    
           & Llama 3-Instruct-Drafter
           \\
            & (314M, draft)
           \\
           \bottomrule
    Layers & 4\\
    Attention heads  & 8\\
    KV heads         & 8 \\ 
    Intermediate dim  & 2816\\
    Hidden dim & 1024\\
    Activation & SiLU
    \end{tabular}
    \label{tab:model_config}
\end{table}

\subsection{Training hyper-parameters}
For draft model pretraining, we used deepspeed stage1 on 32 A100 GPUs. Additionally, during pre-training a large batch-size can be used as compared to distillation which requires the target model weight and output consuming a lot of memory. The optimizer used is \textit{AdamW} with  \textit{WarmUpDecayLR} learning-rate scheduler, maximum learning rate was $0.0001$ while minimum was $1e-6$. 

For draft model fine-tuning, we used deepspeed stage $3$ with batch-size=$40$ on $8$ A-100 GPUs with maximum learning-rate=$0.0003$ with same optimizer and scheduler with $2000$ warm-up steps. 

\subsection{Data processing}
We preprocess the text data with the tokenizer of the target language model while appending End-Of-Sentence (EOS) token at the end of each input sequence. Furthermore, as a postporcessing step, all the sequences are concatenated into chunks of 4096 length, to maximize training throughput without adding pad tokens.



\section{Trimmed draft LM-head}

\subsection{Algorithm Flow}
\label{subsec:appendix_algo}
The VocabTrim follows same inference code as standard SpD systems, with minimal changes related to using a lightweight draft LM-head, and mapping draft model generations back to original vocabulary space as shown in \cref{algorithm:SpD_with_trimmed_vocab}.


\begin{algorithm}
\caption{SpD with \ourmethod{}}
\label{algorithm:SpD_with_trimmed_vocab}
\begin{algorithmic}[1]
\State \textbf{Input:} Prompt $x$, Draft model with trimmed vocab $M_{d\_\text{trim}}$, Target model $M_t$, Draft length $k$, token index mapper $T_{d \rightarrow t}$
\State Initialize output $y \gets x$
\While{not end of sequence}
    \For{$j = 1$ to $k$}
        \State \textcolor{blue}{$\hat{z}_{j} \gets M_{d\_\text{trim}}(y, \hat{y}_{< j})$}
        \State \textcolor{blue}{$\hat{y}_{j} \gets T_{d \rightarrow t}(\hat{z}_{j})$}
    \EndFor
    \State Compute probabilities $P_t \gets M_t(\hat{y}_{1:k})$
    \For{$i = 1$ to $k$}
        \If{$\hat{y}_i$ is consistent with $P_t$}
            \State Append $\hat{y}_i$ to $y$
        \Else
            \State Generate $y_i \gets M_t(y)$
            \State Append $y_i$ to $y$
            \State \textbf{break}
        \EndIf
    \EndFor
\EndWhile
\State \textbf{Return} $y$
\end{algorithmic}
\end{algorithm}

\end{document}